%% file: IEEEMMSP.tex
\documentclass[conference]{IEEEtran}
\IEEEoverridecommandlockouts
\usepackage{cite}
\usepackage{amsmath,amssymb,amsfonts}
\usepackage{algorithmic}
\usepackage{graphicx}
\usepackage{textcomp}
\usepackage{xcolor}

\usepackage{hyperref}
\usepackage{rotating}
\usepackage{overpic}
\usepackage{booktabs} 
\usepackage{multirow,longtable}
\usepackage{colortbl}

\newcommand{\tabincell}[2]{\begin{tabular}{@{}#1@{}}#2\end{tabular}}
\newcommand{\secref}[1]{$\S$ \ref{#1}}
\def\OurModel{\textit{CMA-Net}}
\def\NumOfBaselines{30}

\def\BibTeX{{\rm B\kern-.05em{\sc i\kern-.025em b}\kern-.08em
    T\kern-.1667em\lower.7ex\hbox{E}\kern-.125emX}}
\begin{document}

\title{CMA-Net: A Cascaded Mutual Attention Network for Light Field Salient Object Detection}

\author{\IEEEauthorblockN{Yi Zhang}
\and
\IEEEauthorblockN{Lu Zhang}
\and
\IEEEauthorblockN{Wassim Hamidouche}
\and
\IEEEauthorblockN{Olivier Deforges}
}

\maketitle

\begin{abstract}
In the past few years, numerous deep learning methods have been proposed to address the task of segmenting salient objects from RGB images. However, these approaches depending on single modality fail to achieve the state-of-the-art performance on widely used light field salient object detection (SOD) datasets, which collect large-scale natural images and provide multiple modalities such as multi-view, micro-lens images and depth maps. Most recently proposed light field SOD methods have improved detecting accuracy, yet still predict rough objects' structures and perform slow inference speed. To this end, we propose \OurModel, which consists of two novel cascaded mutual attention modules aiming at fusing the high level features from the modalities of all-in-focus and depth. Our proposed \OurModel~outperforms \NumOfBaselines~state-of-the-art SOD methods on two widely applied light field benchmark datasets. Besides, the proposed \OurModel~is able to inference at the speed of 53 fps. 
Extensive quantitative and qualitative experiments illustrate both the effectiveness and efficiency of our \OurModel, inspiring future development of multi-modal learning for both the RGB-D and light field SOD. 
\end{abstract}


\section{Introduction}\label{sec:intro}

Learning to segment the salient objects that grasp most of the human visual attention from given images is of great importance for various computer vision applications such as video object segmentation \cite{DAVIS} and image caption generation \cite{zhang2019capsal}. In the past years, hundreds of deep learning methods have been proposed to address salient object detection (SOD) in RGB images \cite{zhang2020key}. However, models based on single modality hardly reflect the real human visual mechanism which depends on inputs of multiple modalities (e.g., multi-view images, micro-lens images and depth information), also fail to achieve top performance on challenging multi-modal SOD benchmarks \cite{Zhang2019MemoryorientedDFMoLF}. 

\begin{figure}[t!]
	\centering
    \begin{overpic}[width=0.48\textwidth]{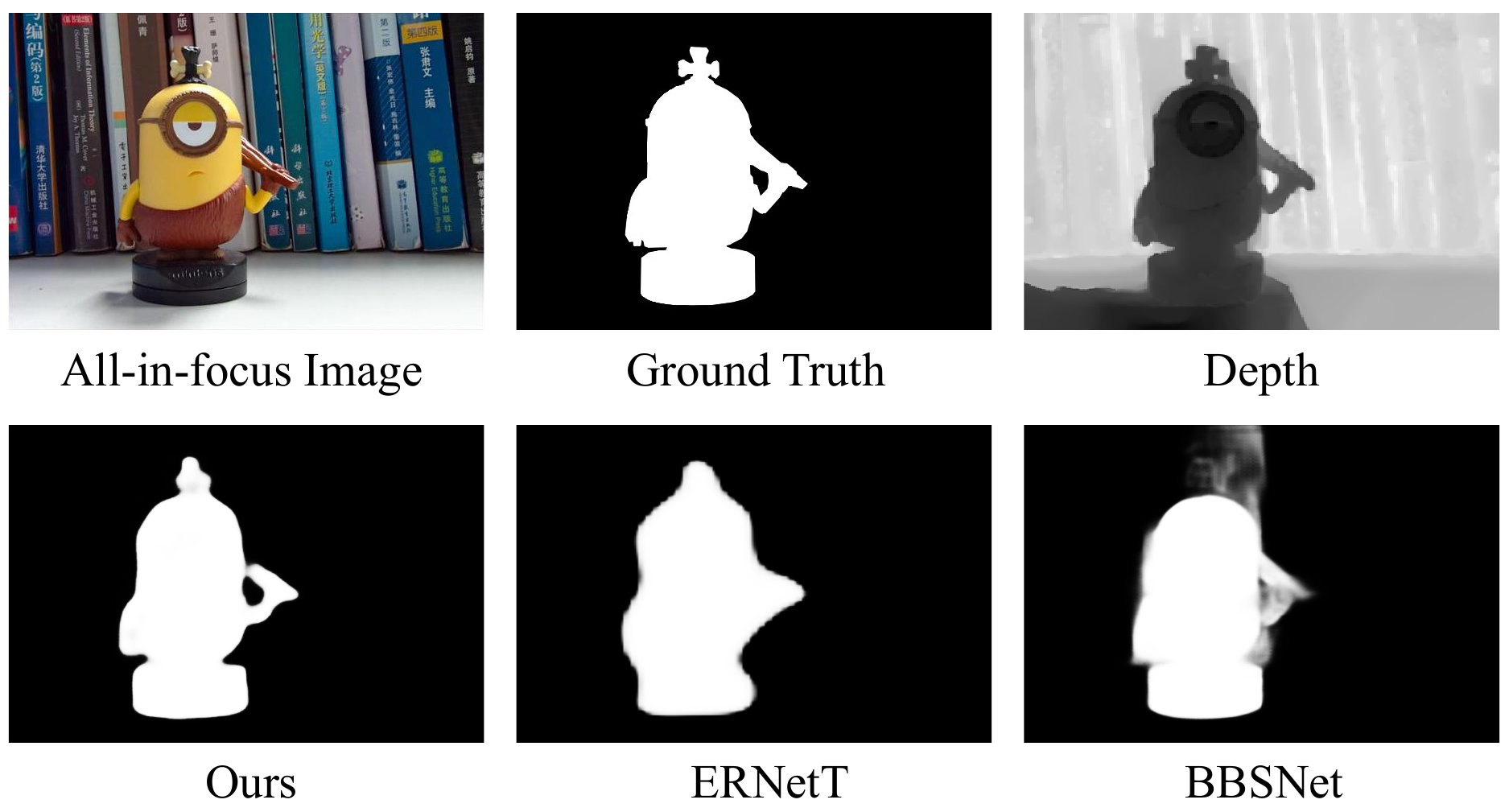}
    \end{overpic}
    \caption{An example of SOD with our \OurModel~and the state-of-the-art multi-modal methods, i.e., ERNetT \protect\cite{Piao2020ExploitARERNet} and BBSNet \protect\cite{fan2020bbs}.}
    \label{fig:examples}
\end{figure}

As the development of light field cameras (e.g. Lytro), it is feasible to collect large-scale natural images in various modalities such as multi-view images, micro-lens images, focal stacks and depth maps \cite{LFSD}. These multi-modal light field data provide rich information to approximate the genuine representation of how light exists in the real world. Recently, SOD on light field appeals increasing attention from the computer vision community, which is attributed to the establishment of light field SOD benchmark datasets including LFSD \cite{LFSD}, HFUT \cite{HFUT}, DUT-LF \cite{DUTLF} and DUT-MV \cite{DUTMV}. To the best of our knowledge, focal stack-based solutions, i.e., ERNet-teacher (ERNetT) \cite{Piao2020ExploitARERNet} and MoLF \cite{Zhang2019MemoryorientedDFMoLF} are the state-of-the-arts on so far the largest light field SOD benchmark \cite{Piao2020ExploitARERNet}, with the training sets of about 1K all-in-focus (AiF) images (also known as RGB images in mono-modal SOD). However, as show in Fig. \ref{fig:examples}, current state-of-the-art RGB-D methods (e.g., BBSNet \cite{fan2020bbs}) or light field approaches (e.g., ERNetT \cite{Piao2020ExploitARERNet}) fail challenging cases by giving poor prediction of salient objects' structures, owning to the inefficiency of traditional attention mechanisms used for multi-modal fusion (details in \secref{subsec:attentions}). Besides, due to the spatial complexity of focal stacks (e.g., a focal stack usually consists of 12 natural images with focuses on multiple depths \cite{Zhang2019MemoryorientedDFMoLF}) corresponding to each of the AiF images, existing top-ranked light field methods (e.g., ERNetT owns an inference speed of 14 fps) are hardly applied to on-line SOD applications.

To this end, we propose a novel mutual attention mechanism established between the modalities of AiF images and depth maps (Fig. \ref{fig:examples}). We further construct a cascaded mutual attention network (\OurModel) to synchronously fuse and decode the high-level features from the two modalities. Our proposed \OurModel~outperforms \NumOfBaselines~SOD methods, also with a relatively high running speed (53 fps). In a nutshell, we provide four main contributions as follows:
\begin{itemize}
    \item We propose a new mutual attention mechanism to efficiently fuse the multi-modal high level features. Note that the proposed attention mechanism can be used as a basic module and embedded to different networks.
    \item We further build \OurModel~which consists of two cascaded novel attention modules, to conduct SOD with the inputs of only two modalities, i.e., AiF images and depth maps. Our \OurModel~ does not apply focal stacks, also avoids processing low-level features from both the modalities, thus performing competitive inference speed.
    \item We conduct systematical benchmark experiments involving \NumOfBaselines~baselines, four metrics and two widely used light field datasets, to illustrate the superiority of our \OurModel.
    \item We present thorough ablation studies to prove the effectiveness of proposed attention mechanism and the necessity of depth information for conducting light field SOD.
\end{itemize}

\section{Related Works}\label{sec:related}

In this section, we briefly introduce recent works from the aspects of light field SOD datasets, methods, and attention mechanisms used for cross modal fusion.

\subsection{Light Field SOD}\label{sec:related_datasets}
\subsubsection{Methods} To the best of our knowledge, the existing methods include 10 traditional ones (\cite{LFSD,Li2015WSC,Sheng2016RelativeLFRL,Wang2017BIF,HFUT,Wang2018AccurateSSDDF,Wang2018SalienceGDSGDC,Wang2020RegionbasedDRDFD,Piao2020SaliencyDVDCA,Zhang2015SaliencyDILF}) and 7 deep learning-based ones (\cite{DUTLF,DUTMV,Zhang2020LFNetLFNet,Lytro,zhang2020multiMTCNet,Zhang2019MemoryorientedDFMoLF,Piao2020ExploitARERNet}), respectively. Traditional methods such as DILF \cite{Zhang2015SaliencyDILF}, BIF \cite{Wang2017BIF}, MA \cite{HFUT}, SGDC \cite{Wang2018SalienceGDSGDC} and DCA \cite{Piao2020SaliencyDVDCA} used depth maps as auxiliary inputs, and took advantage of handcrafted features such as color contrast and objects' location cues towards multi-modal SOD.
Methods such as ERNetT \cite{Piao2020ExploitARERNet} and MoLF \cite{Zhang2019MemoryorientedDFMoLF} applied stacked natural images in various depths, thus achieving good performances on light field benchmarks. However, these methods tend to be slow during testing, due to the high computational cost induced by the inputs consisting of focal stacks.

\subsection{Attention for Cross Modal Fusion}\label{subsec:attentions}
In the field of multi-modal SOD, several cross modal attention mechanisms have been recently proposed. S2MA \cite{LiuS2MA} designed a self-mutual attention module to automatically select useful high-level features learned from both modalities. BBSNet \cite{fan2020bbs} constructed depth-enhanced module to combine the channel and spatial attention at each of the encoding stages. DLLF \cite{DUTLF}, LFNet \cite{Zhang2020LFNetLFNet}, MoLF \cite{Zhang2019MemoryorientedDFMoLF} and ERNetT \cite{Piao2020ExploitARERNet} all employed classical channel attention \cite{hu2018squeeze} to aid the feature selection and refinement from the modality of focal stacks. Recent large-scale light field SOD benchmark studies (e.g., \cite{Zhang2019MemoryorientedDFMoLF,Piao2020ExploitARERNet}) indicate that it remains un open question how to efficiently fuse the intrinsic features from multiple modalities for advanced detecting accuracy.

\begin{figure*}[t!]
	\centering
    \begin{overpic}[width=0.87\textwidth]{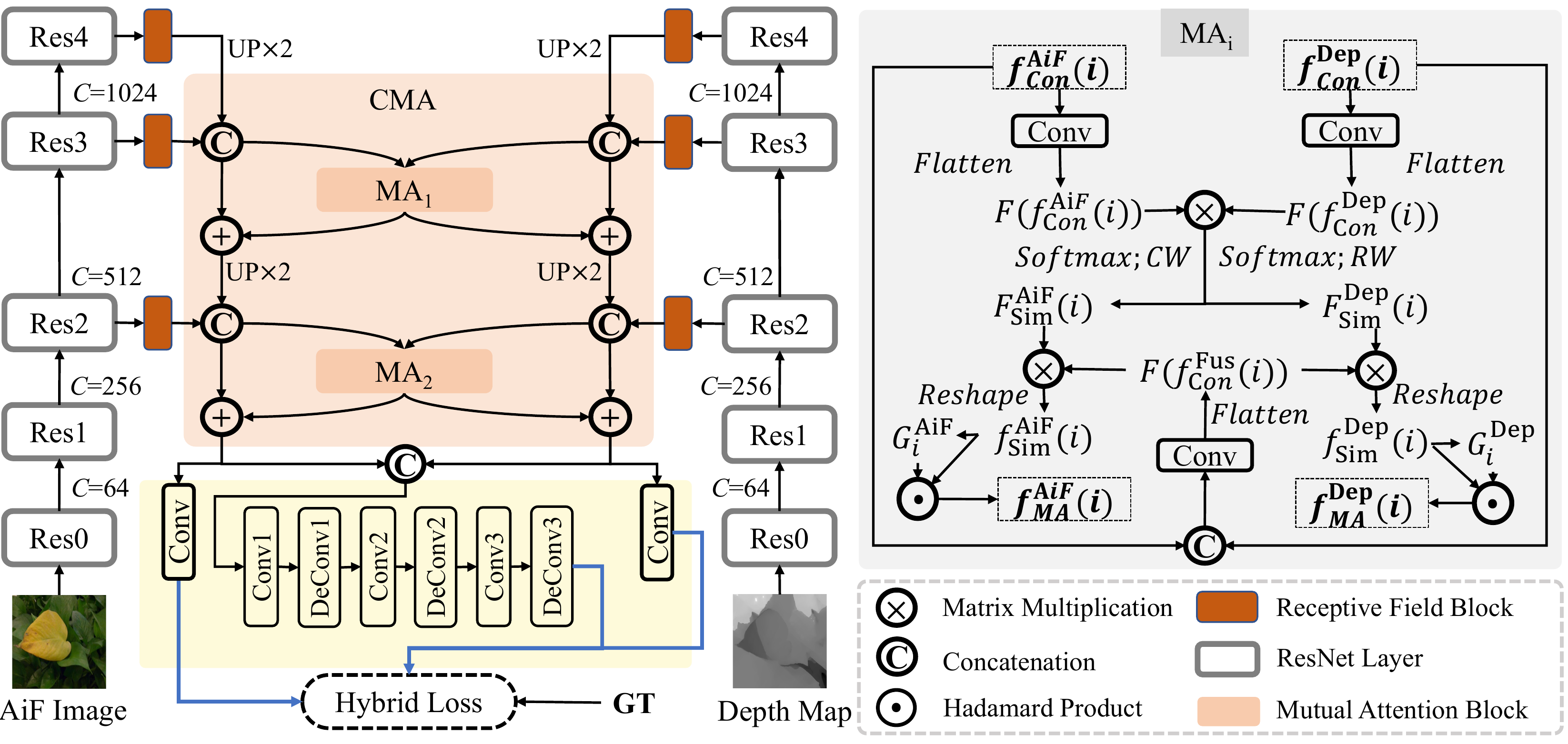}
    \end{overpic}
	\caption{An overview of our \textbf{\OurModel}. RGB-D high level features extracted from duel-branch encoder are fed into two proposed cascaded mutual attention modules, followed by a group of (de-)convolutional layers from \cite{fan2020bbs}. The abbreviations in the figure are detailed as follows: AiF Image = all-in-focus image. GT = ground truth. Res$i$ = the $i$th ResNet \cite{he2016deep} layer. (De)Conv = (de-)convolutional layer. MA$_{i}$ = the $i$th mutual attention module. CMA = cascaded mutual attention module. CW = column-wise normalization. RW = row-wise normalization.}
    \label{fig:model}
\end{figure*}

\section{Cascaded Mutual Attention Network}\label{sec:model}

The \OurModel~consists of a duel-branch ResNet50 \cite{he2016deep}-based encoder and a cascaded mutual attention-based decoder.

\subsection{RGB-D Encoder}
Our encoder is a duel-branch architecture that consists of symmetrical convolutional layers transferred from ImageNet-pretrained ResNet50 \cite{he2016deep}. In \OurModel, we only process the high-level features, i.e., the features ($\{f_{i}^{\text{AiF}}\}_{i=2}^{4}$ and $\{f_{i}^{\text{Dep}}\}_{i=2}^{4}$) from the last three layers of ResNets, to focus on salient objects' shape and location cues \cite{wu2019cascaded} also to avoid extra computational cost. The $\{f_{i}^{\text{AiF}}\}_{i=2}^{4}$ and $\{f_{i}^{\text{Dep}}\}_{i=2}^{4}$ are then fed into a series of receptive field blocks \cite{wu2019cascaded} to enrich the global context information from each encoding level (Fig. \ref{fig:model}).

\subsection{Cascaded Mutual Attention}
\subsubsection{Multi-level Concatenation}
The refined high level features from adjacent encoding stages, e.g., $f_{\text{RFB}}^{\text{AiF}}(i)$ and $f_{\text{RFB}}^{\text{AiF}}(i+1)$ are further concatenated as $f_{\text{Con}}^{\text{AiF}}(i)$, where $i$ ($i \in \{2,3\}$) denotes the $i$th decoding stage corresponding to the $i$th ResNet layer. 
\subsubsection{Mutual Attention}
Continually taking the $i$th decoding stage as an example, a similarity matrix ($Sim_i$) between the features from two branches is computed as:
\begin{equation}\label{equ:similarity}
   Sim_i = F(f_{\text{Con}}^{\text{Dep}}(i))^{\text{T}} \otimes F(f_{\text{Con}}^{\text{AiF}}(i)),
\end{equation}
where $F(\cdot)$ represents a flatten operation reshaping the 3D feature matrix
$f_{\text{Con}}^{\text{AiF}}(i) \in \mathbb{R}^{H\times W\times C}$
to a 2D one with a dimension of $C \times HW$, $\otimes$ denotes matrix multiplication. Inspired by \cite{lu2020zero}, the $Sim_i$ is then column-/row-wisely normalized via:
\begin{equation}\label{equ:normalization}
\begin{split}
   F_{\text{Sim}}^{\text{AiF}}(i) &= Softmax(Sim_i) \in [0, 1]^{{HW \times HW}},\\
   F_{\text{Sim}}^{\text{Dep}}(i) &= Softmax(Sim_i^{\text{T}}) \in [0, 1]^{{HW \times HW}},
 \end{split}
\end{equation}
where $Softmax(\cdot)$ normalizes each column of the $HW \times HW$ matrix. As shown in Fig. \ref{fig:model}, the mutual attentions ($f_{\text{Sim}}^{\text{AiF}}(i)$, $f_{\text{Sim}}^{\text{Dep}}(i)$) for each of the branches are computed as:
\begin{equation}\label{equ:attention_corse}
\begin{split}
  f_{\text{Sim}}^{\text{AiF}}(i) &= R(F_{\text{Sim}}^{\text{AiF}}(i) \otimes  F(f_{\text{Con}}^{\text{Fus}}(i))^{\text{T}}) \in [0, 1]^{{H \times W \times C}},\\
  f_{\text{Sim}}^{\text{Dep}}(i) &= R(F_{\text{Sim}}^{\text{Dep}}(i) \otimes F(f_{\text{Con}}^{\text{Fus}}(i))^{\text{T}}) \in [0, 1]^{{H \times W \times C}},
   \end{split}
\end{equation}
where $R(\cdot)$ reshapes the given matrix from a dimension of $C \times HW$ to $H \times W \times C$, $F(f_{\text{Con}}^{\text{Fus}}(i))$ denotes fused features from both branches (Fig. \ref{fig:model}), which is the main difference when compared to \cite{zhang2021learning}. To further avoid unstable feature updating during the model training process, a pair of self-adapted gate functions ($G_i^{AiF}$, $G_i^{Dep}$) are computed to gain the final mutual attention matrix ($f_{\text{MA}}^{\text{AIF}}(i)$, $f_{\text{MA}}^{\text{Dep}}(i)$). The process can be described as:
\begin{equation}\label{equ:attention_refined}
   f_{\text{MA}}^{\text{AIF}}(i) = G_i^{AiF} \odot  f_{\text{Sim}}^{\text{AiF}}(i)~\text{and}~f_{\text{MA}}^{\text{Dep}}(i) = G_i^{Dep} \odot f_{\text{Sim}}^{\text{Dep}}(i),
\end{equation}
where $\odot$ represents Hadamard product, the gate function $G_i^{AiF} = \sigma(Conv(f_{\text{Sim}}^{\text{AiF}}(i)))$ with $Conv(\cdot)$ and $\sigma(\cdot)$ denoting a convolutional layer and a Sigmoid function, respectively. In \OurModel, we cascade two identical mutual attention modules to establish the decoder, thus acquiring the best performance (see detailed ablation studies in \secref{sec:ablation}).

\subsection{Co-Supervision and Hybrid Loss}
As shown in Fig. \ref{fig:model}, to stabilize the multi-modal learning process, we apply a three-way strategy to co-supervise the training of our \OurModel. Besides, inspired by a multi-loss function training setting applied in \cite{wei2020f3net}, we combine three loss functions including widely used binary cross entropy loss ($\ell_{\text{BCE}}$), intersection over union loss ($\ell_{\text{IoU}}$) and E-loss ($\ell_{\text{EM}} = 1 - E_\phi$), which is based on a recently proposed SOD metric ($E_\phi$ \cite{Fan2018Enhanced}). Therefore, our hybrid loss function is denoted as:
\begin{equation}\label{equ:loss}
   \ell = \sum_{n=1}^{N}\ell_{\text{BCE}}(P_{n}, G) + \ell_{\text{IoU}}(P_{n}, G) + \ell_{\text{EM}}(P_{n}, G),
\end{equation}
where $\{P_n\}_{n=1}^{3}\in [0,1]$ denote the predicted three-way saliency maps, while $G \in \{0,1\}$ denotes the corresponding ground-truth binary mask.

\subsection{Implementation Details}
Our \OurModel~is implemented in PyTorch 1.8 and optimized with Adam algorithm \cite{kingma2014adam}. During the training stage, the batch size is set to 16, the learning rate is initialized as 1e-4 with a decay rate of 0.1 for every 50 epochs. It takes about one hour to finish the training of \OurModel~based on a platform consists of Intel$^\circledR$ Xeon(R) W-2255 CPU @ 3.70GHz and one Quadro RTX 6000 GPU.

\begin{table}[t]
	\centering
	\caption{Quantitative results for different models on two benchmark datasets. The best scores are in \textbf{boldface}. We train and test our \OurModel~with the settings that are consistent with \protect\cite{Piao2020ExploitARERNet}, which is the state-of-the-art model at present. $\star$ indicates tradition methods. $\uparrow$ indicates the higher the score the better, and vice versa for $\downarrow$.}
	\label{tab:quantitative}
	\renewcommand{\arraystretch}{1.5}
	\setlength\tabcolsep{4pt}
	\resizebox{0.49\textwidth}{!}{
		\begin{tabular}{l||cccc||cccc}
			\hline
			\toprule
			 \multirow{2}{*}{{Models}} 
			& \multicolumn{4}{c||}{\tabincell{c}{DUT-LF~\cite{DUTLF}}} 
			& \multicolumn{4}{c}{\tabincell{c}{HFUT~\cite{HFUT}}} \\
			\cline{2-9}
			& $F_\beta\uparrow$ & $S_\alpha\uparrow$ & $E_\phi\uparrow$ & $M\downarrow$
			& $F_\beta\uparrow$ & $S_\alpha\uparrow$ & $E_\phi\uparrow$ & $M\downarrow$
			\\
			\hline
			\input{Figures/table_data}
			\bottomrule
		\end{tabular}
	}
\end{table}

\begin{figure}[t!]
	\centering
	\begin{overpic}[width=0.47\textwidth]{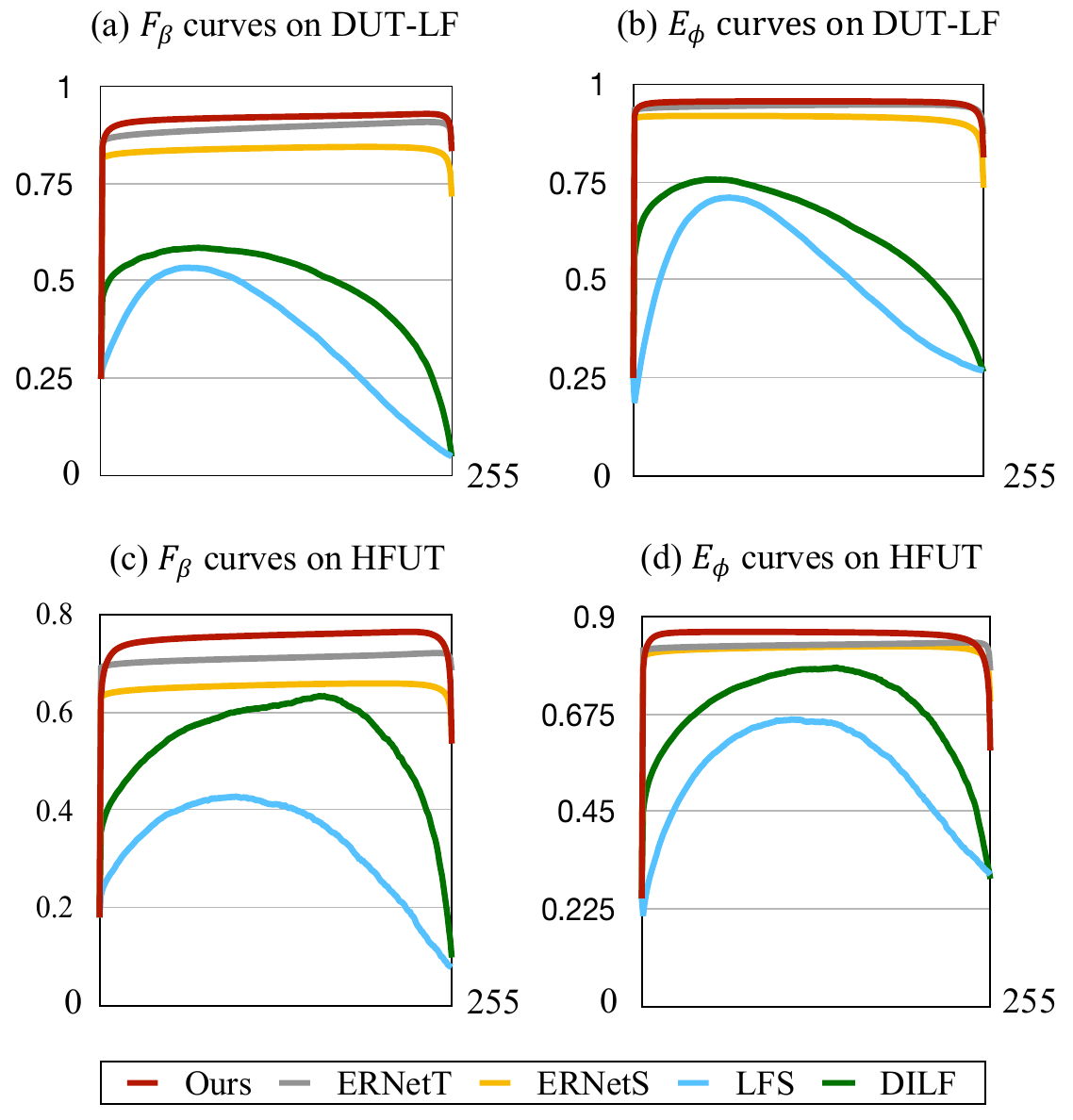}
    \end{overpic}
	\caption{F-measure ($F_\beta$) and E-measure ($E_\phi$) curves of state-of-the-art light field SOD models and our \OurModel~upon two benchmark datasets.}
    \label{fig:curves}
\end{figure}

\begin{figure*}[t!]
	\centering
    \begin{overpic}[width=0.99\textwidth]{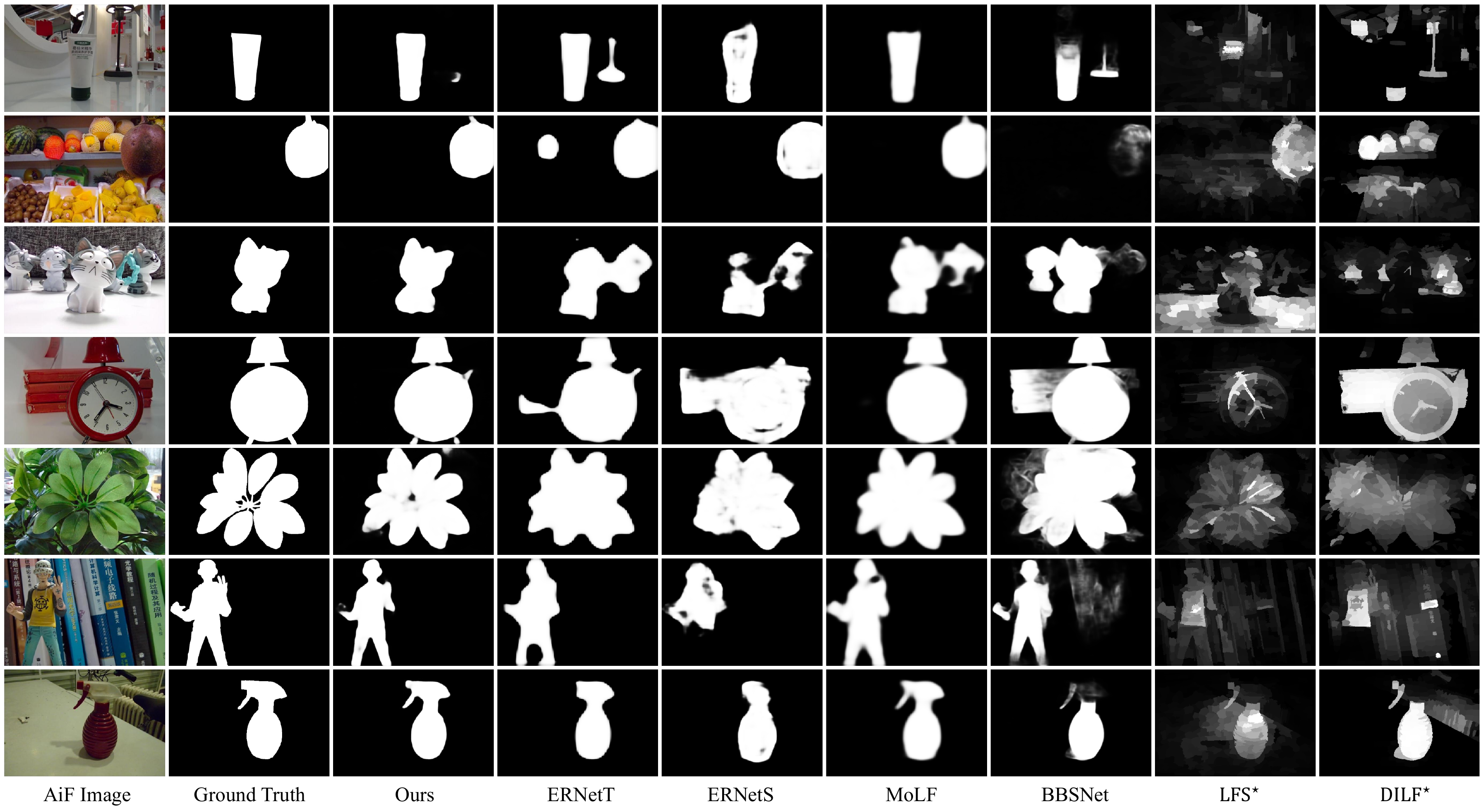}
    \end{overpic}
	\caption{Visual results our \OurModel~and state-of-the-art multi-modal SOD models. $\star$ denotes traditional methods. AiF Image = all-in-focus image.}
    \label{fig:qualitative}
\end{figure*}

\section{Experiments}\label{sec:experiments}

\subsection{Datasets and Evaluations}
We evaluate our \OurModel~and \NumOfBaselines~baselines on two widely used light field SOD datasets, i.e., DUT-LF \cite{DUTLF} and HFUT \cite{HFUT} (more description in \secref{sec:related_datasets}). We follow the training/testing settings in \cite{Piao2020ExploitARERNet}, which is so far the largest light field SOD benchmark.

To quantitatively compare our \OurModel~with the others, we apply four widely applied SOD metrics, including recently proposed S-measure ($S_\alpha$) \cite{fan2017structure} and E-measure ($E_\phi$) \cite{Fan2018Enhanced}, generally agreed mean absolute error ($M$) \cite{perazzi2012saliency} and F-measure ($F_\beta$) \cite{achanta2009frequency}. Specifically, $S_\alpha$ evaluates the structure similarities between salient objects in ground-truth maps and predicted saliency maps:
\begin{equation}\label{equ:sm}
   S_\alpha = \alpha \times S_{o} + (1 - \alpha) \times S_{r}.
\end{equation}
where $S_{o}$ and $S_{r}$ denotes the object-/region-based structure similarities, respectively. $\alpha \in [0,1]$ is set as 0.5 so that equal weights are assigned to both the object-level and region-level assessments \cite{fan2017structure}. $E_\phi$ is a cognitive vision-inspired metric to evaluate both the local and global similarities between two binary maps. It is defined as:
\begin{equation}\label{equ:em}
   E_{\phi}=\frac{1}{W\times H}\sum_{x=1}^W\sum_{y=1}^H\phi\left(P(x,y), G(x,y)\right),
\end{equation} 
where $\phi$ represents the enhanced alignment matrix \cite{Fan2018Enhanced}, $H$ and $W$ denotes height and width, respectively. $M$ computes the mean absolute error between the ground truth $G \in \{0, 1\}$ and a normalized predicted saliency map $P \in [0, 1]$, i.e.,
\begin{equation}\label{equ:mae}
   M = \frac{1}{W\times{H}}\sum_{i=1}^{W}\sum_{j=1}^{H}\mid G(i, j) - P(i, j)\mid,
\end{equation}
$F_\beta$ gives a single-valued metric by considering both the $Precision$ and $Recall$:
\begin{equation}\label{equ:fm}
   F_{\beta} = \frac{(1+\beta^{2})Precision \times Recall}{\beta^{2}Precision + Recall}, 
\end{equation}
with
\begin{equation}
  Precision=\frac{\left|F\cap G\right|}{\left|F\right|}; Recall=\frac{\left|F\cap G\right|}{\left|G\right|},
\end{equation}
where $F$ denotes a binary mask converted from a predicted saliency map and $G$ is the ground truth. Multiple $F$ are computed by taking different thresholds of $[0,255]$ on the predicted saliency map. The $\beta^{\text{2}}$ is set to 0.3 according to \cite{achanta2009frequency}. In this paper, we report adaptive $F_\beta$/$E_\phi$, thus being consistent with ERNetT \cite{Piao2020ExploitARERNet} and MoLF \cite{Zhang2019MemoryorientedDFMoLF}.

\subsection{Comparison with the State-of-the-Arts}

\subsubsection{Quantitative Results}
As shown in Tab. \ref{tab:quantitative}, our \OurModel~outperforms \NumOfBaselines~state-of-the-art methods on all four metrics. Besides, as shown in Fig. \ref{fig:curves}, our \OurModel~is also able to acquire the best performance on each threshold ($\tau, \tau \in [0,255]$) of F/E-measure curves. The sufficient quantitative results indicate the effectiveness as well as robustness of the proposed \OurModel.

\subsubsection{Qualitative Results}
Some visual results of our \OurModel~and six advanced baselines are shown in Fig. \ref{fig:qualitative}. The \OurModel~predicts saliency maps that are closest to the ground-truth, by finely depicting the shapes and boundaries of the labeled salient objects.

\subsubsection{Inference Speed}
It is worth mentioning that our \OurModel~is capable of running at 53 fps, being much more efficient than the top-ranked ERNetT \cite{Piao2020ExploitARERNet} which owns an inference speed of only 14 fps.

\subsection{Ablation Studies}\label{sec:ablation}
We conduct thorough ablation studies to further verify the effectiveness of each module of the proposed method. We first construct basic model-1 which consists of single-branch ResNet layers and a group of (de-)convolutional layers, without the inputs of depth maps. Followed by model-2, which contains the duel-branch encoder with both AiF images and depth maps as inputs. As a result, we find that depth information can be helpful for the SOD task (Tab. \ref{tab:ab_quantitative}). We then carefully add mutual attention mechanisms to different decoding stages. The model-3 and model-4 are embedded with one mutual attention module at the $2^{nd}$ and $3^{rd}$ stages (\secref{sec:model}), respectively. Finally, we cascade two mutual attention modules (\OurModel) and thus gaining the best performance compared to all ablated versions (Tab. \ref{tab:ab_quantitative}).

\begin{table}[t]
	\centering
	\caption{Quantitative results for the ablation studies of \OurModel~on DUT-LF~\protect\cite{DUTLF} and HFUT~\protect\cite{HFUT}. The best scores are in \textbf{boldface}. $\uparrow$ indicates the higher the score the better, and vice versa for $\downarrow$.}
	\label{tab:ab_quantitative}
	\renewcommand{\arraystretch}{1.2}
	\setlength\tabcolsep{2pt}
	\resizebox{0.42\textwidth}{!}{
		\begin{tabular}{cc||ccccc}
			\hline
			\toprule
			\multicolumn{2}{c||}{Metric} & Model-1 & Model-2 & Model-3 & Model-4 & \OurModel \\
			\midrule
			 \multirow{4}{*}{DUT-LF} &  \multirow{1}{*}{$F_\beta\uparrow$} & \multirow{1}{*}{0.879} & \multirow{1}{*}{0.895} & \multirow{1}{*}{0.914} & \multirow{1}{*}{0.915} & \multirow{1}{*}{\textbf{0.917}}  \\
			 &  \multirow{1}{*}{$S_\alpha\uparrow$} & \multirow{1}{*}{0.893} & \multirow{1}{*}{0.911} & \multirow{1}{*}{0.916} & \multirow{1}{*}{0.916} & \multirow{1}{*}{\textbf{0.918}} \\
			&  \multirow{1}{*}{$E_\phi\uparrow$} & \multirow{1}{*}{0.931} & \multirow{1}{*}{0.943} & \multirow{1}{*}{0.949} & \multirow{1}{*}{\textbf{0.950}} & \multirow{1}{*}{0.949} \\
			&  \multirow{1}{*}{$M\downarrow$} & \multirow{1}{*}{0.047} & \multirow{1}{*}{0.039} & \multirow{1}{*}{0.034} & \multirow{1}{*}{0.034} & \multirow{1}{*}{\textbf{0.033}} \\
			 	\midrule
			   	 \multirow{4}{*}{HFUT} &  \multirow{1}{*}{$F_\beta\uparrow$} & \multirow{1}{*}{0.697} & \multirow{1}{*}{0.704} & \multirow{1}{*}{0.727} & \multirow{1}{*}{0.729} & \multirow{1}{*}{\textbf{0.744}}  \\
			 &  \multirow{1}{*}{$S_\alpha\uparrow$} & \multirow{1}{*}{0.792} & \multirow{1}{*}{0.795} & \multirow{1}{*}{0.791} & \multirow{1}{*}{0.791} & \multirow{1}{*}{\textbf{0.807}} \\
			&  \multirow{1}{*}{$E_\phi\uparrow$} & \multirow{1}{*}{0.837} & \multirow{1}{*}{0.828} & \multirow{1}{*}{0.842} & \multirow{1}{*}{0.858} & \multirow{1}{*}{\textbf{0.865}} \\
			&  \multirow{1}{*}{$M\downarrow$} & \multirow{1}{*}{0.074} & \multirow{1}{*}{0.078} & \multirow{1}{*}{0.076} & \multirow{1}{*}{0.071} & \multirow{1}{*}{\textbf{0.069}} \\
		
			\bottomrule
		\end{tabular}
	}
\end{table}

\section{Conclusion}\label{sec:conclusion}
In this paper, we conduct light field SOD by proposing a new deep learning method, \OurModel, which consists of two cascaded novel mutual attention modules for RGB-D cross modal high-level feature fusion. Our \OurModel~achieves the best performance on widely used light field benchmark datasets based on four commonly agreed SOD metrics, also with a competing inference speed of 53 fps. Systematical ablation studies are also conducted to verify the effectiveness of the proposed attention mechanism as well as the feasibility of RGB-D cross modal learning strategy for light field SOD.

\bibliographystyle{IEEEMMSP}
\bibliography{IEEEMMSP}

\end{document}

%% file: Figures/table_data.tex
	
    	Ours	&	\textbf{0.917}	&	\textbf{0.918}	&	\textbf{0.949}	&	\textbf{0.033}	&	\textbf{0.744}	&	\textbf{0.807}	&	\textbf{0.865}	&	\textbf{0.069}		\\
		ERNetT\cite{Piao2020ExploitARERNet}	&	0.889	&	0.899	&	0.943	&	0.040	&	0.705	&	0.777	&	0.831	&	0.082 \\
		ERNetS\cite{Piao2020ExploitARERNet}	&	0.838	&	0.848	&	0.916	&	0.061	&	0.651	&	0.736	&	0.824	&	0.085 \\
		MoLF\cite{Zhang2019MemoryorientedDFMoLF}	&	0.843	&	0.887	&	0.923	&	0.052	&	0.627	&	0.742	&	0.785	&	0.095 \\
		DLFS\cite{DUTMV}	&	0.801	&	0.841	&	0.891	&	0.076	&	0.615	&	0.741	&	0.783	&	0.098 \\
		DILF$^{\star}$\cite{Zhang2015SaliencyDILF}	&	0.641	&	0.705	&	0.805	&	0.168	&	0.555	&	0.695	&	0.736	&	0.131 \\
		LFS$^{\star}$\cite{LFSD}	&	0.484	&	0.563	&	0.728	&	0.240	&	0.430	&	0.579	&	0.686	&	0.205 \\

     BBSNet\cite{fan2020bbs} & 0.854 & 0.865 & 0.908 & 0.066 & 0.705 & 0.783 & 0.829 & 0.078  \\
    UCNet\cite{zhang2020uc} & 0.817 & 0.831 & 0.878 & 0.081 & 0.710 & 0.770 & 0.830 & 0.084  \\
    S2MA\cite{LiuS2MA} & 0.754 & 0.787 & 0.841 & 0.103 & 0.647 & 0.761 & 0.787 & 0.100  \\
    D3Net\cite{fan2020rethinking} & 0.790 & 0.822 & 0.869 & 0.084 & 0.692 & 0.778 & 0.827 & 0.080 \\
   	CPFP\cite{zhao2019contrast}	&	0.730	&	0.741	&	0.808	&	0.101	&	0.594	&	0.701	&	0.768	&	0.096 \\
		TANet\cite{Chen2019TANet}	&	0.771	&	0.803	&	0.861	&	0.096	&	0.638	&	0.744	&	0.789	&	0.096 \\
		MMCI\cite{Chen2019MMCI}	&	0.750	&	0.785	&	0.853	&	0.116	&	0.645	&	0.741	&	0.787	&	0.104 \\
		PDNet\cite{zhu2019pdnet}	&	0.763	&	0.803	&	0.864	&	0.111	&	0.629	&	0.770	&	0.786	&	0.105 \\
		PCA\cite{chen2018progressively}	&	0.762	&	0.800	&	0.857	&	0.100	&	0.644	&	0.748	&	0.782	&	0.095 \\
		CTMF\cite{Han2017CTMF}	&	0.790	&	0.823	&	0.881	&	0.100	&	0.620	&	0.752	&	0.784	&	0.103 \\
		DF\cite{qu2017rgbd}	&	0.733	&	0.716	&	0.838	&	0.151	&	0.562	&	0.670	&	0.742	&	0.138 \\
\hline	

     F3Net\cite{wei2020f3net} & 0.882 & 0.888 & 0.900 & 0.057 & 0.718 & 0.777 & 0.815 & 0.095 \\
    GCPANet\cite{GCPANet} & 0.867 & 0.885 & 0.898 &  0.064 & 0.691 & 0.777 & 0.799 & 0.105 \\ 
   	EGNet\cite{zhao2019egnet}	&	0.870	&	0.886	&	0.914	&	0.053	&	0.672	&	0.772	&	0.794	&	0.094 \\
		CPD\cite{wu2019cascaded}	&	0.887	&	0.890	&	0.923	&	0.050	&	0.689	&	0.764	&	0.810	&	0.097 \\
		PoolNet\cite{liu2019simple}	&	0.868	&	0.889	&	0.919	&	0.051	&	0.683	&	0.776	&	0.802	&	0.092 \\
		PAGRN\cite{zhang2018progressive}	&	0.828	&	0.822	&	0.878	&	0.084	&	0.635	&	0.717	&	0.773	&	0.114 \\
		C2S\cite{li2018contour}	&	0.791	&	0.844	&	0.874	&	0.084	&	0.650	&	0.763	&	0.786	&	0.111 \\
		R$^{3}$Net\cite{deng2018r3net}	&	0.783	&	0.819	&	0.833	&	0.113	&	0.625	&	0.727	&	0.728	&	0.151 \\
		Amulet\cite{zhang2017amulet}	&	0.805	&	0.847	&	0.882	&	0.083	&	0.636	&	0.767	&	0.760	&	0.110 \\
		UCF\cite{zhang2017learning}	&	0.769	&	0.837	&	0.850	&	0.107	&	0.623	&	0.754	&	0.764	&	0.130 \\
		SRM\cite{wang2017stagewise}	&	0.832	&	0.848	&	0.899	&	0.072	&	0.672	&	0.762	&	0.801	&	0.096 \\
		NLDF\cite{luo2017non}	&	0.778	&	0.786	&	0.862	&	0.103	&	0.636	&	0.729	&	0.807	&	0.091 \\
		DSS\cite{hou2017deeply}	&	0.728	&	0.764	&	0.827	&	0.128	&	0.626	&	0.715	&	0.778	&	0.133 \\